\documentclass[10pt,twocolumn,letterpaper]{article}

\usepackage{cvpr}
\usepackage{times}
\usepackage{epsfig}
\usepackage{graphicx}
\usepackage{amsmath}
\usepackage{amssymb}
\usepackage{caption}
\usepackage{bm}
\usepackage[pagebackref=true,breaklinks=true,letterpaper=true,colorlinks,bookmarks=false]{hyperref}
\usepackage{gensymb}

\usepackage[breaklinks=true,bookmarks=false]{hyperref}

\cvprfinalcopy 

\def\cvprPaperID{5480} 

\begin{document}

\title{Rotate-and-Render: Unsupervised Photorealistic Face Rotation \\ from Single-View Images}

\author{Hang Zhou$^{1}$\thanks{Equal Contribution.} \quad Jihao Liu$^{2}$\footnotemark[1] \quad Ziwei Liu$^1$ \quad  Yu Liu$^{1}$\thanks{Corresponding author.} \quad Xiaogang Wang$^{1}$ \\
    $^1$The Chinese University of Hong Kong \quad
    $^2$SenseTime Research\\
    {\tt\small \{zhouhang@link,yuliu@ee,xgwang@ee\}.cuhk.edu.hk}\hspace{6pt}
    {\tt\small liujihao@sensetime.com}\hspace{6pt}
    {\tt\small zwliu.hust@gmail.com}
    \vspace{-10pt}
}

\twocolumn[{
\renewcommand\twocolumn[1][]{#1}%
\maketitle
\vspace{-17pt}
\begin{center}
 \centering
 \includegraphics[width=0.95\textwidth]{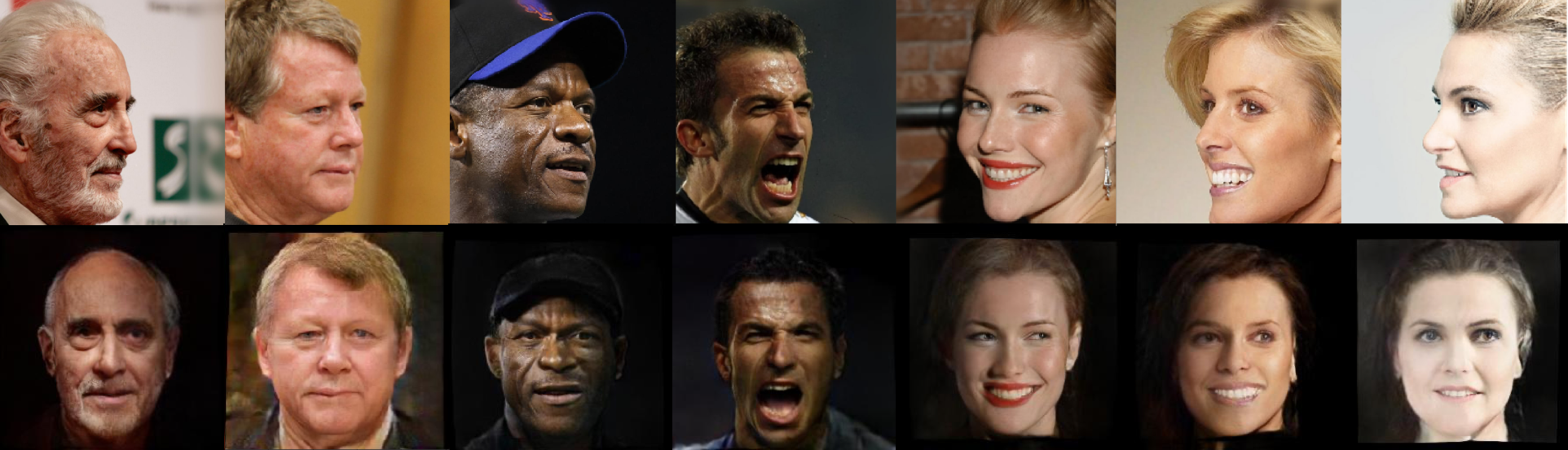}
 \captionof{figure}{\textbf{The rotated faces synthesized by our approach on CelebA-HQ dataset.} The first row is the input and the second row is our result. It can be observed that our unsupervised framework produces near photo-realistic results even under extreme poses and expressions. 
 }
 \label{fig:celeba4}
\end{center}
}]

{
  \renewcommand{\thefootnote}%
    {\fnsymbol{footnote}}
  \footnotetext[1]{Equal contribution.}
  \footnotetext[2]{Corresponding author.}
}

\maketitle

\begin{abstract}
Though face rotation has achieved rapid progress in recent years, the lack of high-quality paired training data remains a great hurdle for existing methods.
The current generative models heavily rely on datasets with multi-view images of the same person. 
Thus, their generated results are restricted by the scale and domain of the data source.
To overcome these challenges, we propose a novel unsupervised framework that can synthesize photo-realistic rotated faces using only single-view image collections in the wild.
Our key insight is that \textbf{rotating} faces in the 3D space back and forth, and re-\textbf{rendering} them to the 2D plane can serve as a strong self-supervision. 
We leverage the recent advances in 3D face modeling and high-resolution GAN to constitute our building blocks.
Since the 3D rotation-and-render on faces can be applied to arbitrary angles without losing details, our approach is extremely suitable for in-the-wild scenarios (\ie no paired data are available), where existing methods fall short.
Extensive experiments demonstrate that our approach has superior synthesis quality as well as identity preservation over the state-of-the-art methods, across a wide range of poses and domains.
Furthermore, we validate that our rotate-and-render framework naturally can act as an effective data augmentation engine for boosting modern face recognition systems even on strong baseline models\footnote{Code and models are available at: \url{https://github.com/Hangz-nju-cuhk/Rotate-and-Render}.}.

\end{abstract}

\section{Introduction}

Face rotation, or more generally speaking multi-view face synthesis, has long been a topic of great research interests, due to its wide applications in computer graphics, augmented reality and particularly, face recognition. It is also an ill-posed task with inherent ambiguity that can not be well solved by existing methods.
Traditionally, this problem is addressed by using 3D models such as 3DMM~\cite{blanz1999morphable}. A common challenge here is that invisible areas would appear when rotating a 3D-fitted face. Previous researchers propose to solve this problem through symmetric editing and invisible region filling~\cite{zhu2015high}. However, this filling process usually introduces visible artifacts which lead to non-realistic results.

With the rapid progress of deep learning and generative adversarial networks (GANs), reconstruction-based methods have been widely applied to face frontalization and rotation~\cite{yim2015rotating,huang2017beyond,tran2017disentangled,hu2018pose,tian2018cr}. 
Due to the information loss when encoding the given face to bottleneck embeddings, reconstruction-based methods often suffer from the loss of local details such as known facial textures and shapes. It further leads to the confusion of identity information. 
Moreover, the most notable drawback of existing reconstruction-based methods is that multi-view data of the same person has to be provided as direct supervisions in most cases. To this end, the datasets used for training are constraint to ones in controlled environments such as Multi-PIE~\cite{gross2010multi}, and synthetic ones such as 300W-LP~\cite{zhu2015high}. 
Models trained on controlled datasets can only generate results within a specific domain, lacking the desired generalization ability. Also, their generated resolutions are normally limited to under $128 \times 128$, far from perceptually satisfying.

To overcome these challenges, we propose a novel unsupervised framework that can synthesize photorealistic rotated faces using only single-view image collections in the wild, and can achieve arbitrary-angle face rotation.
While details and ID information tend to degrade during the encoding process of 2D-based methods, we propose to keep as much known information about the given face as possible with 3D model. 

\begin{figure}[t!]
\centering
\includegraphics[width=1\linewidth]{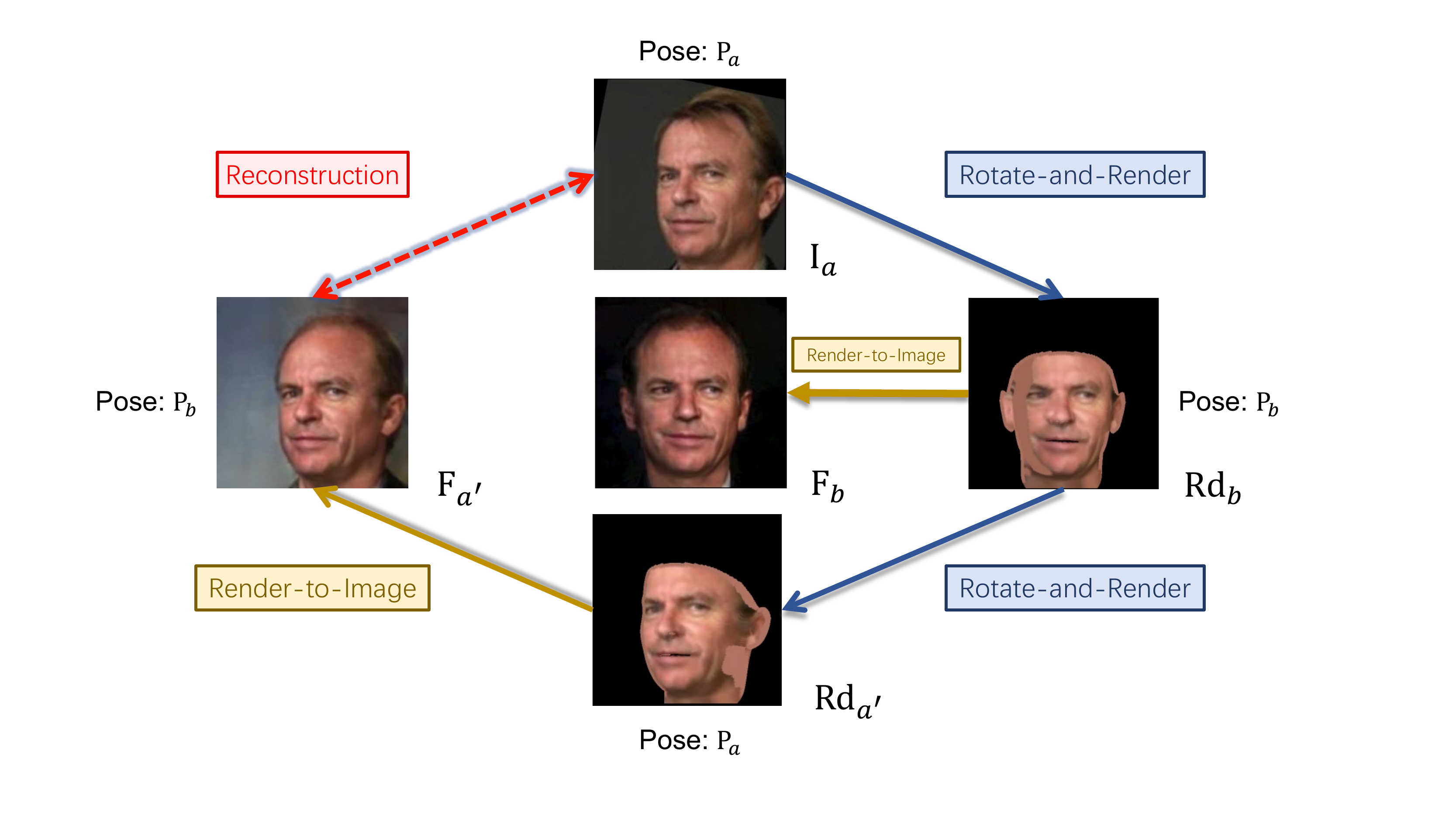}
\caption{\textbf{Overview of our unsupervised face rotation framework from only single-view image collections.} We \textbf{rotate} the 3D-mesh firstly to an arbitrary pose $\text{P}_b$, and \textbf{render} it to the 2D space to get a 2D-rendered image $\text{Rd}_b$. Then we \textbf{rotate} it back to its original position $\text{P}_a$ and \textbf{render} it to be $\text{Rd}_{a'}$ using textures extracted from $\text{Rd}_b$. Finally we use an render-to-image translation network to fill the invisible parts and map the rendered image to real image domain.}
\label{fig:problem}
\vspace{-5pt}
\end{figure}

\emph{Our key insight is that \textbf{rotating} faces in the 3D space back and forth, and re-\textbf{rendering} them to the 2D plane can serve as a strong self-supervision.} 
We take the advantage of both 3D face modeling and GANs by using off-the-shelf 3D-fitting network 3DDFA~\cite{zhu2017face} and the neural renderer~\cite{kato2018neural}. Invisible parts would appear to one fixed 2D view when a face is rotated from one pose to another. While previous methods require both images to form an \{input, ground truth\} pair, we use the self-supervision of one single image. The key is to create and then eliminate the artifacts caused by rotations. Given one face at pose $\text{P}_a$, we \textbf{rotate} its 3D-mesh firstly to another arbitrary pose $\text{P}_b$, and \textbf{render} it to the 2D space to get a 2D-rendered image $\text{Rd}_b$. Then we \textbf{rotate} it back to its original position and \textbf{render} it to be $\text{Rd}_{a'}$ using textures extracted from $\text{Rd}_b$. Finally, we use an image-to-image translation network to fill the invisible parts and map the rendered image to real image domain. The overview of our pipeline is shown in Fig.~\ref{fig:problem}. In this way, existing local texture information can be preserved while GAN is responsible for fixing the occluded parts. As the whole pipeline rotates and renders a face forward and backward, we term it \textbf{Rotate-and-Render} framework.

Remarkably, our proposed framework does not rely on paired data or any kind of label, thus any face image can be used as our training source. With unlimited training data, our model can be leveraged to boost large-scale face recognition, providing augmentations and alignments for profile faces. 
While previous methods are often evaluated on small datasets with moderate baselines, we validate the effectiveness of our approach for large-scale face recognition on strong baseline models.

Our contributions are summarized as follows: 
{\bf 1)} We propose a novel Rotate-and-Render framework for training face rotation in a fully unsupervised manner under in-the-wild scenarios. No paired data or any label is needed. 
{\bf 2)} We convert incomplete rendered images to real images using an image-to-image translation network, with which photo-realistic face rotation results can be generated. 
{\bf 3)} We validate that our generation results benefit large-scale face recognition even on strong baseline models.

\section{Related Work}

\subsection{Face Rotation and Multi-View Synthesis}
The problem of face rotation aims to synthesize multi-view faces given a single face image regardless of its viewpoint. Among all views, the frontal view particularly attracts much more research interests. 
Traditionally, this problem is tackled by building 3D models and warping textures on 3D or 2D~\cite{hassner2015effective,zhu2015high}. OpenGL can also be used to also easier cases~\cite{moniz2018unsupervised}.
However, their synthesized results are usually blurry and not photorealistic.

\noindent\textbf{Reconstruction-based Methods.}
Recently, with the progress of deep learning~\cite{liu2015deep} and GANs~\cite{goodfellow2014generative}, reconstruction-based models have revolutionized the field of face frontalization~\cite{tran2017disentangled,huang2017beyond,hu2018pose,tran2018representation,qian2019make,qian2019unsupervised}. DR-GAN~\cite{tran2017disentangled,tran2018representation}, for the first time, adopts GAN to generate frontal faces with an encoder-decoder architecture. Although they do not use multi-view data, the generated results are not satisfying and have perceptually-visible artifacts. Then TP-GAN~\cite{huang2017beyond} utilizes global and local networks together with a multi-task learning strategy to frontalize faces. CAPG-GAN~\cite{hu2018pose} uses face heatmap to guide the generation of multi-view faces. 
Most of the methods are trained on Multi-PIE dataset, which makes them overfit the dataset's environment and cannot generalize well to unseen data. Even FNM~\cite{qian2019unsupervised} which proposes to combine both labeled and unlabeled data can only normalize faces to a standard MultiPIE-like frontal-view. 
The common shortcoming that almost all of them share is the requirement of paired multi-view training data.

\noindent\textbf{3D Geometry-based Methods.}
Several attempts have been made to incorporate 3D prior knowledge into GAN-based frontalization pipeline. FF-GAN~\cite{yin2017face} proposes to integrate the regression of 3DMM coefficients into the network and employ them for generation. But its generation results are of low-quality. UV-GAN~\cite{deng2018uv} proposes to complete UV-map using image-to-image translation, which is similar to our work. However, their pipeline requires high precision 3D fitting and ground-truth UV-maps, which are both difficult to obtain. Besides, their generated results are not photorealistic under in-the-wild scenarios. Recently, HF-PIM~\cite{cao2018learning} achieves high-quality face frontalization results using facial texture map and correspondence fields. However, their method also requires paired data for training.
In this work, our proposed approach not only gets rid of the requirement for paired training data, but also has the capacity to generate high-quality results which preserve texture and identity information.

\subsection{Image-to-Image Translation}
Image-to-image translation aims at translating an input image to a corresponding output image, typically from a different modality or domain. The core idea is to predict pixel values directly with encoder-decoder architecture and low-level connections. GANs are widely adopted in this field~\cite{pix2pix2017,CycleGAN2017,wang2018high,park2019semantic,wang2018image,ge2018fd,zhou2019talking,yin2019instance,liu2019learning,lee2019maskgan}, since the adversarial loss can alleviate the blurry issue by $L_1$ reconstruction loss. Pix2Pix framework~\cite{pix2pix2017} firstly uses image-conditional GANs for this task. Pix2PixHD~\cite{wang2018high} used stacked structures to produce high-quality images. Recently, SPADE~\cite{park2019semantic} and MaskGAN~\cite{lee2019maskgan} discovers that semantic information can be infused via conditional batch normalization to further improve the results.

\noindent\textbf{Cycle Consistency in Unsupervised Image-to-Image Translation.} Cycle consistency has been proven useful on various tasks~\cite{CycleGAN2017,pumarola2018ganimation,wayne2018reenactgan,wang2019learning}, particularly for image translation without paired data. For example, CycleGAN~\cite{CycleGAN2017} achieves unpaired image-to-image translation, and GANimation~\cite{pumarola2018ganimation} proposes to generation animation without supervision. Our idea shares similar intuition with cycle consistency. However, The difference is that for most unpaired translation papers, they focus on mapping across domain by training neural networks,
while our proposed Rotate-and-Render operation is performed off-line.

\noindent\textbf{Inpainting with GAN.} After the rendering step, our target is to synthesize photorealistic images from renders with artifacts, which is substantially a combination of image translation and inpainting. Recently, advanced inpainting techniques also benefit from the image-to-image translation pipeline with GAN~\cite{pathak2016context,yeh2017semantic,wang2018image,zhou2019vision}. Therefore we adopt the recent advances in image-to-image translation to realize render-to-image generation.

\begin{figure}[t!]
\centering
\includegraphics[width=1\linewidth]{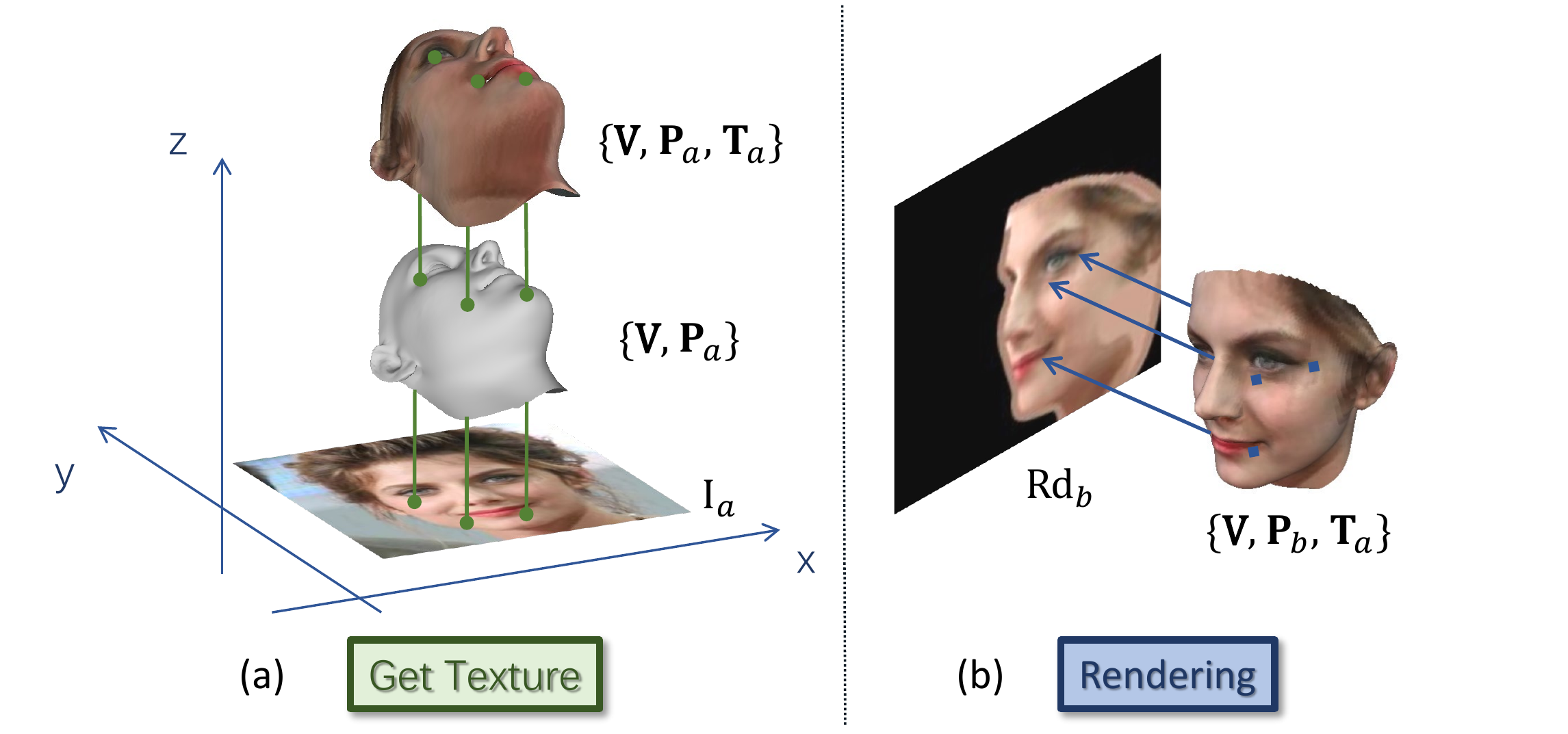}

\caption{The process of getting texture and rendering. 3D points lying on the same lines will correspond to the same texture in the 2D space. The subscript $a$ and $b$ are associated with pose $a$ and $b$ in Fig~\ref{fig:pipeline}.}

\label{fig:texture}
\vspace{-5pt}
\end{figure}



\begin{figure*}[t!]
\centering
\includegraphics[width=1\linewidth]{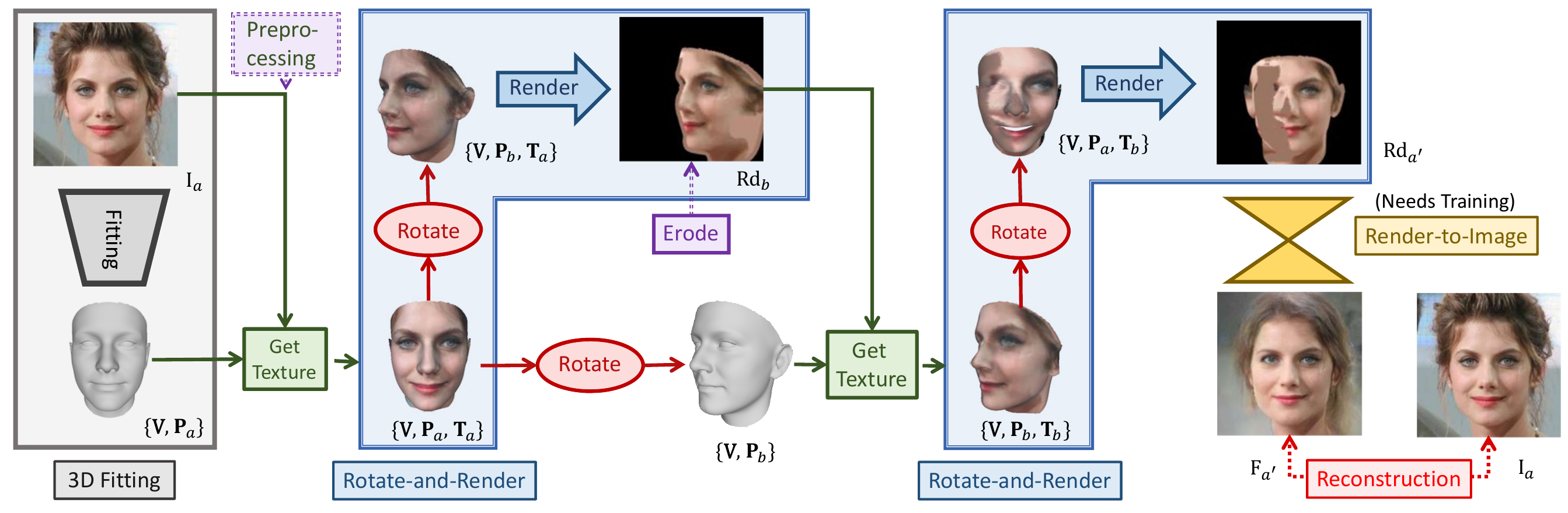}

\caption{\textbf{Our framework for unsupervised photorealistic face rotation.} Our key insight is that \textbf{rotating} faces in the 3D space back and forth, and re-\textbf{rendering} them to the 2D plane can serve as a strong self-supervision. After two rotate-and-render processes, we can create a rendered image $\text{Rd}_{a'}$ which has the artifacts of a face rotated to the position $a$ from any position $b$. So that the input image $\text{I}_a$ itself can serve as ground truth for training. Only the Render-to-Image module needs training during the whole process.}
\label{fig:pipeline}
\vspace{-5pt}
\end{figure*}


\section{Our Approach}

\noindent\textbf{Overall Framework.}
Our whole face rotation framework consists of three parts: 3D face fitting, the rotate-and-render strategy for training data preparation, and the render-to-image translation module. We elaborate each component as follows.

\subsection{3D Face Modeling and Render}
\noindent{\textbf{3D Face Fitting.}}  Our method relies on a rough 3D parametric fitting of a given face, where all kinds of 3D models and predicting methods are applicable. Here we briefly introduce the notations and concepts of 3D face model we use as an instruction for the following sections. 

Given one definition of 3D face model with $n$ vertices, the shape prediction of a face ${\bf{V}} = [\bm{v}_1, \bm{v}_2, \cdots, \bm{v}_n]$ represents the normalized position of its vertices in the 3D space with $\bm{v}_i = [x_i, y_i, z_i]^{\text{T}}$.
The projection of a 3D shape onto the 2D image can be written as:
\begin{align}
    \label{eq1}
    \Pi({\bf{V}}, {\bf{P}}) = f * {\bf{p_r}} * {\bf{R}} * {\bf{V}} + {\bf{h_{2d}}},
\end{align}
where $\Pi$ is the projection function that maps model vertices to their 2D positions. The matrix multiplication is denoted by ``$*$''. $f$ is the scale factor, ${\bf{p_r}}$ is the orthographic projection matrix, ${\bf{R}}$ is the rotation matrix and ${\bf{h_{2d}}}$ is the 2D shift. Here we regard all the above defined projection related parameters to be a joint representation of the face's relative pose  ${\bf{P}} = \{f, {\bf{R}}, {\bf{h_{2d}}}\}$. 
\\

\noindent{\textbf{Acquiring Textures.}} Textures also play a crucial role when transforming a complete 3D representation to 2D space. For each vertex $\bm{v}_i$, there exists an associated texture $\bm{t}_i = [r_i, g_i, b_i]^\text{T}$ on a colored face. We use the simplest vertical projection to get the colors of the vertices from the original image ${\text{I}}$. The color of each vertex can be written as :
\begin{align}
    \label{eq2}
    \bm{t}_i = {\text{I}}(\Pi(\bm{v}_i, {\bf{P}})),
\end{align}
where $\Pi(\bm{v}_i, {\bf{P}})$ is the projected 2D coordinate of the vertice $\bm{v}_i$. In this way we get all corresponding textures $\textbf{T} = [\bm{t}_1, \cdots, \bm{t}_n]$. 
This process can be easily depicted in Fig~\ref{fig:texture}. We refer to the whole process of getting textures uniformly as:
\begin{align}
    \label{eq3}
    {\bf{T}} = \text{GetTex}(\text{I}, \{{\bf{V}}, {\bf{P}}\}).
\end{align}

The projected result $\Pi({\bf{V}, {\bf{P}}})$ is irrelevant to the vertices' $z$ coordinate due to the orthographic matrix ${\bf{p_r}}$. For each position $(x_j, y_j)$ on the 2D space, there might exist multiple rotated vertices on the line $\{x = x_j$ and $y = y_j\}$ in 3D space, then the same texture will be assigned to every one of them. For all $\bm{v}_k\in \{\bm{v}~|~(x_j, y_j) =  \Pi(\bm{v}_k, {\bf{P}})\}$, only the outermost one with the largest $z$ axis value gets the correct texture. Its index is 
\begin{align}
    \label{eq4}
    K_{j} = \mathop{\arg\max}_{k} ([0, 0, 1] * \bm{v}_k).
\end{align}
The rest are actually invisible vertices due to occlusion in the 2D space. We keep the wrongly acquired textures and regard them as artifacts that we aim to deal with.
\\

\noindent{\textbf{Rendering.}} Given a set of 3D representation of a face $\{{\bf{V}}, {\bf{P}}, {\bf{T}}\}$, rendering is to map it to the 2D space and generate an image. The rendering process is the reverse of acquiring texture as depicted in Fig~\ref{fig:texture} (b). Same as equation~\ref{eq4}, it is known that $K_j$ is the index of outermost vertice given a 2D point $(x_j, y_j)$. The rendered color image $\text{Rd}$ can be calculated as:
\begin{align}
    \label{eq5}
   \text{Rd}(x_j, y_j) =\left\{
\begin{aligned}
{\bf{T}}\{K_j\}, ~~~\exists K_j \in \mathbb{N},\\
0, 
~~~~~~~~~~~\nexists K_j \in \mathbb{N}.
\end{aligned}
\right.
\end{align}
Finally, we denote the whole rendering process as:
\begin{align}
    \label{eq6}
     \text{Rd} = \text{Render}(\{{\bf{V}}, {\bf{P}}, {\bf{T}}\} ).
\end{align}
We use the open-sourced Neural Mesh Renderer~\cite{kato2018neural} to perform rendering without any training.


\subsection{Rotate-and-Render Training Strategy}
It can be discovered that with an accurately fitted 3D model, our aim is to fill the invisible vertices with the correct textures for getting another view of a face. However, existing works with similar ideas~\cite{cao2018learning,deng2018uv} require ground truth supervision from multi-view images which are difficult to get. 

Here we propose a simple strategy to create training pairs called Rotate-and-Render (R$\&$R) which consists of two rotate-and-render operations. The key idea is to create the artifacts caused by rotating occluded facial surface to the front and eliminate them. Thus we can leverage only self-supervision to train networks. The whole pipeline and visualizations of each step are all illustrated in Fig~\ref{fig:pipeline}. 

Given an input image $\text{I}_a$, we firstly get the 3D model parameters by a 3D-face fitting model:
\begin{align}
    \label{eq7}
    \{{\bf{V}}, {\bf{P}}_a\} = \text{Fitting}(\text{I}_a),
\end{align}
where $a$ denotes the current view of this face in the 2D space, with ${\bf{P}}_a = \{ f, {\bf{R}}_a, {\bf{h_{2d}}} \}$. The textures of its vertices can be acquired as:
\begin{align}
    \label{eq8}
    {\bf{T}}_a = \text{GetTex}(\text{I}_a, \{{\bf{V}}, {\bf{P}}_a\}).
\end{align}
We then rotate the 3D representation of this face to another random 2D view $b$ by multiplying ${\bf{R}}_a$ with another rotation matrix ${\bf{R}}_{random}$ to get ${\bf{P}}_b = \{ f, {\bf{R}}_b, {\bf{h_{2d}}} \}$. And we render the current 3D presentation to $\text{Rd}_b = \text{Render}(\{{\bf{V}}, {\bf{P}}_b, {\bf{T}}_a\})$. This completes the first rotate-and-render operation.

Under this circumstance, another set of textures can be acquired as:
\begin{align}
    \label{eq9}
   {\bf{T}}_b = \text{GetTex}(\text{Rd}_b, \{{\bf{V}}, {\bf{P}}_b\}).
\end{align}
%
We discover that the vertice set whose textures are correct under view $b$ is the subset of the vertice set whose textures are correct under view $a$. So unlike previous works rely on a ground truth image $\text{I}_b$ as supervision to recover a face under view $b$ given view $a$, we propose to recover $\textbf{T}_a$ regarding $\textbf{T}_b$ as input.

Specifically, we rotate the 3D position ${\bf{P}}_b$ back to $ {\bf{P}}_a$ and render it back to its original 2D position with
\begin{align}
    \label{eq10}
   \text{Rd}_{a'} = \text{Render}(\{{\bf{V}}, {\bf{P}}_a, {\bf{T}}_b\}).
\end{align}
This $\text{Rd}_{a'}$ is basically a rendered image with artifacts caused by rotating a face from view $b$ to $a$ in the 2D space. In this way, we get our input/ground truth pair $\{\text{Rd}_{a'}, \text{I}_a\}$ for training.


\begin{figure}[t!]
\centering
\includegraphics[width=1\linewidth]{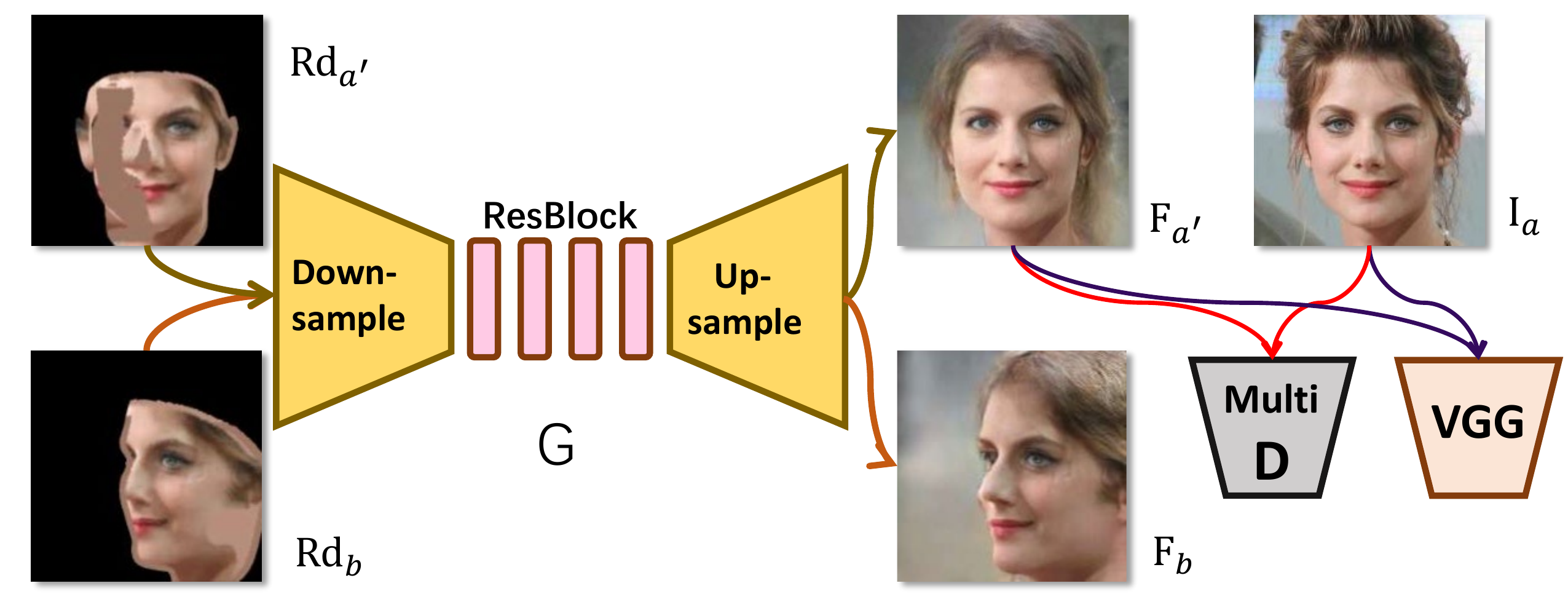}

\caption{Our render-to-image module. Rendered images $\text{Rd}_{a'}$ and $\text{Rd}_b$ are sent into the generator \text{G} to get the results $\text{F}_{a'}$ and $\text{F}_b$. Losses are conducted between generated $\text{F}_{a'}$ and the ground truth $\text{I}_a$ through the discriminator and the pretrained VGG network.}

\label{fig:render2image}
\vspace{-5pt}
\end{figure}


\subsection{Render-to-Image Generation}
In order to eliminate the artifacts and map the rendered images $\text{Rd}_b$ and $\text{Rd}_{a'}$ from the rendered domain to real image domain, we propose the render-to-image generation module to create $\text{F}_{a'} = \text{G}(\text{Rd}_{a'})$ and $\text{F}_{b} = \text{G}(\text{Rd}_b)$ using generator $\text{G}$, as shown in Fig~\ref{fig:render2image}.

The basic generator $\text{G}$ is adopted from CycleGAN~\cite{CycleGAN2017}, which is enough to handle most images in our datasets. The multi-layer discriminator and perceptual loss from Pix2PixHD~\cite{wang2018high} are borrowed directly. The loss function of the discriminator includes the adversarial loss
\begin{align}
    \label{eq11}
    {\mathcal{L}}_{\text{GAN}}{(\text{G}, \text{D})} =\mathbb{E}_{\text{I}}[\log \text{D}(\text{I}_a)]\ + \mathbb{E}_{\text{Rd}}[\log (1 - \text{D}(\text{G}(\text{Rd}_{a'}))],
\end{align}
and a feature matching loss. The feature matching loss is realized by extracting features from multiple layers of the discriminator and regularizing the distance between input and generated images. We use $F^{(i)}_{D}(\text{I})$ to denote the feature extracted from the $i$th layer of the discriminator for an input $\text{I}$. For total $N_D$ layers, the feature matching loss can be written as:
\begin{align}
    \label{eq12}
    {\mathcal{L}}_{FM}{(G, D)} = \frac{1}{N_D} \sum_{i=1}^{N_D}{\|F^{(i)}_{D}(\text{I}_a) - F^{(i)}_{D}(\text{G}(\text{Rd}_{a'})) \|_1 }.
\end{align}
Perceptual loss is achieved by using ImageNet pretrained VGG network. It is used to regularize both the generation results and the generated identity. The perceptual loss is very similar to that of $\mathcal{L}_{FM}$, with the features denoted by $F^{(i)}_{vgg}$, the loss function is:
\begin{align}
    \label{eq13}
    {\mathcal{L}}_{vgg}{(\text{G}, \text{D})} = \frac{1}{N_{vgg}} \sum_{i=1}^{N_{vgg}}{\|F^{(i)}_{vgg}(\text{I}_a) - F^{(i)}_{vgg}(\text{G}(\text{Rd}_{a'})) \|_1 }.
\end{align}
Our full objective function can be written as:
\begin{align}
{\mathcal {L}}_{total} = {\mathcal{L}}_{\text{GAN}} + \lambda_1\mathcal{L}_{FM} +  \lambda_2{\mathcal{L}}_{vgg}.
\end{align}

During testing, our desired output can be directly generated by assigning the target view ${\bf{P}_c}$ to form $\text{Rd}_c = \text{Render}(\{{\bf{V}}, {\bf{P}}_c, {\bf{T}}_a\})$, and send the rendered ${\text{Rd}_c}$ into the trained generator $\text{G}$.


\begin{figure}[t!]
\centering
\includegraphics[width=1\linewidth]{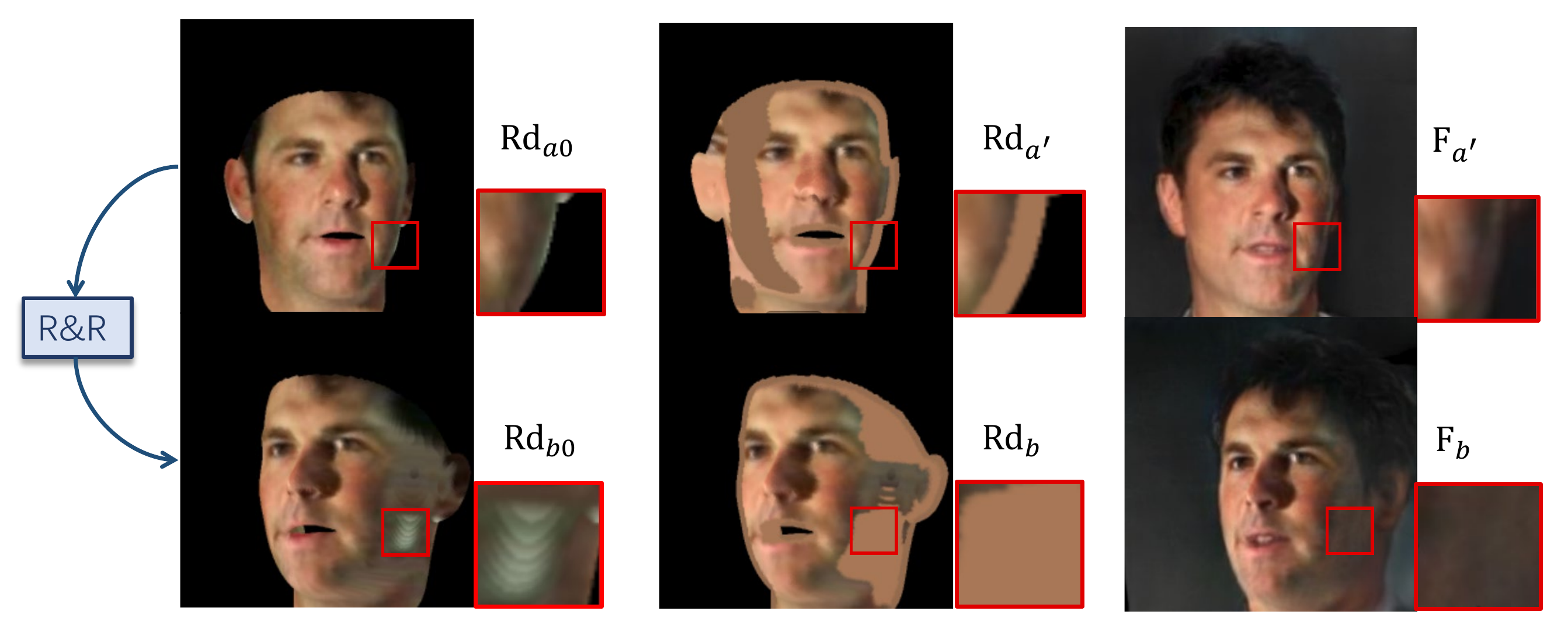}

\caption{Illustration of the miss-alignment effect and the results of the erosion preprocessing.}

\label{fig:erode}
\vspace{-5pt}
\end{figure}


\subsection{Building Block Details}

\noindent{\textbf{3D Fitting Tool.}} Our choice of 3D face model is the 3D Morpohable Model (3DMM)~\cite{blanz1999morphable}. Its flattened vertice matrix ${\bf{V}}$ can be denoted by ${\bf{S}} = [\bm{v}_1^\text{T}, \bm{v}_2^\text{T}, \cdots, \bm{v}_n^\text{T}]^\text{T}$. Its description of 3D faces is based on PCA: 
\begin{align}
    \label{eq15}
    {\bf{S}} = {\bf{\overline{S}}} + {\bf{A}}_{id}{\bm{\alpha}}_{id} + {\bf{A}}_{exp}{\bm{\alpha}}_{exp}.
\end{align}
Here ${\bf{\overline{S}}}$ is the mean shape, ${\bf{A}}_{id}$ and ${\bf{A}}_{exp}$ are the principle axes for identities and expressions, respectively.

We use the open-sourced implementation and the pretrained model of 3DDFA~\cite{zhu2017face} for 3D face fitting. It is a deep learning based model for regressing the parameters $[{\bf{P}}^\text{T}, {\bm{\alpha}}_{id}, {\bm{\alpha}}_{exp}]$, from a single 2D face image. Therefore the face's 3D shape and its relative position in 2D space $\{{\bf{V}}, {\bf{P}}\}$ can be predicted. Please be noted that we do not train the 3DDFA. The usage of it is only for fast application, and its miss-alignment on 3D shape prediction can cast certain problems to the final results. An alternative way is to tune the 3D model using identity labels. However, in a fully unsupervised setting, we propose to solve it using a novel eroding preprocessing. 
\\

\noindent{\textbf{Pre-Render and Erosion.}} Due to the inaccuracy of 3D fitting methods, the projection of wrongly fitted vertices sometimes lies outside the true edges of a face. When this kind of 3D miss alignment happens, background pixels would be assigned to these vertices. During the rotate-and-render process, those wrongly acquired textures will be rendered onto the rotated $\text{Rd}_b$ (see Fig.~\ref{fig:erode} left column). However, such artifacts are difficult to create directly on $\text{Rd}_{a'}$ by the rotate-and-render process, which means that they do not exist in training input-output pairs. Thus they cannot be handled by our generator.

The way to solve it is to pre-render the fitted 3D representation $\{\textbf{V}, \textbf{P}_a, \textbf{T}_a\}$ to $\text{Rd}_{a0}$, and erode the rendered image by certain pixels. The erosion is performed basing on the projected edge of ${\bf{V}}$ with an average color of all vertices. Then texture $\textbf{T}_a$ is renewed to 
\begin{align}
    \label{eq16}
{\bf{T}}_a = \text{GetTex}(\text{erode}({\text{Rd}}_a), \{{\bf{V}}, {\bf{P}}_a\}).
\end{align}
So that $\text{Rd}_{b}$ can only contain artifacts that exist in $\text{Rd}_{a'}$. The output after the erosion can be found at Fig.~\ref{fig:erode}.

\section{Experiments}

\subsection{Experimental Settings}

\noindent{\textbf{Implementation Details}}. 
We firstly run 3DDFA across all datasets to get the parameters $\{\bf{V}, \bf{P}\}$ for all images. 
With known $\bf{V}$, we are able to know the facial key points and perform alignment for faces according to their eye centers and noses. 
The generator $\text{G}$ contains 4 downsample and upsample blocks and 9 residual blocks. Spectral Normalization~\cite{miyato2018spectral} and Batch Normalization~\cite{ioffe2015batch} are applied to all layers. The discriminator consists of two scales.

Our models are trained using Pytorch on 8 Tesla V100 GPUs with 16 GB memory. Two graphical cards are used for rendering and the others for training. 
The time for rotating one single image is about 0.1s. 
The weights $\lambda_1$ and $\lambda_2$ are both set to 10. 
Please refer to our code and models for more details.

\noindent{\textbf{Datasets.}} Using our rotate-and-render strategy, we do not rely on any paired multi-view data or supervision, so there is no problem of over-fitting in evaluation. 
Theoretically, we have unlimited number of data for training our system. As datasets in controlled environments are not always applicable to real-world scenarios, we focus more on large-scale in-the-wild problems.

\textbf{CASIA-WebFace}~\cite{yi2014learning} and \textbf{MS-Celeb-1M}~\cite{guo2016ms} are selected as training sets for both our render-to-image module and our face recognition network. Specifically, we adopt the version named \textbf{MS1MV2} cleaned in~\cite{deng2019arcface}.
For evaluating our boost on strong face recognition systems, we test on the standard \textbf{LFW}~\cite{huang2008labeled}, \textbf{IJBA}~\cite{klare2015pushing} which contain profile faces in videos, and \textbf{MegaFace 1-million Challenge}~\cite{nech2017level}. MegaFace is the most challenging and wildly applied dataset for evaluating face recognition results.

\begin{figure}[t!]
\centering
\includegraphics[width=1\linewidth]{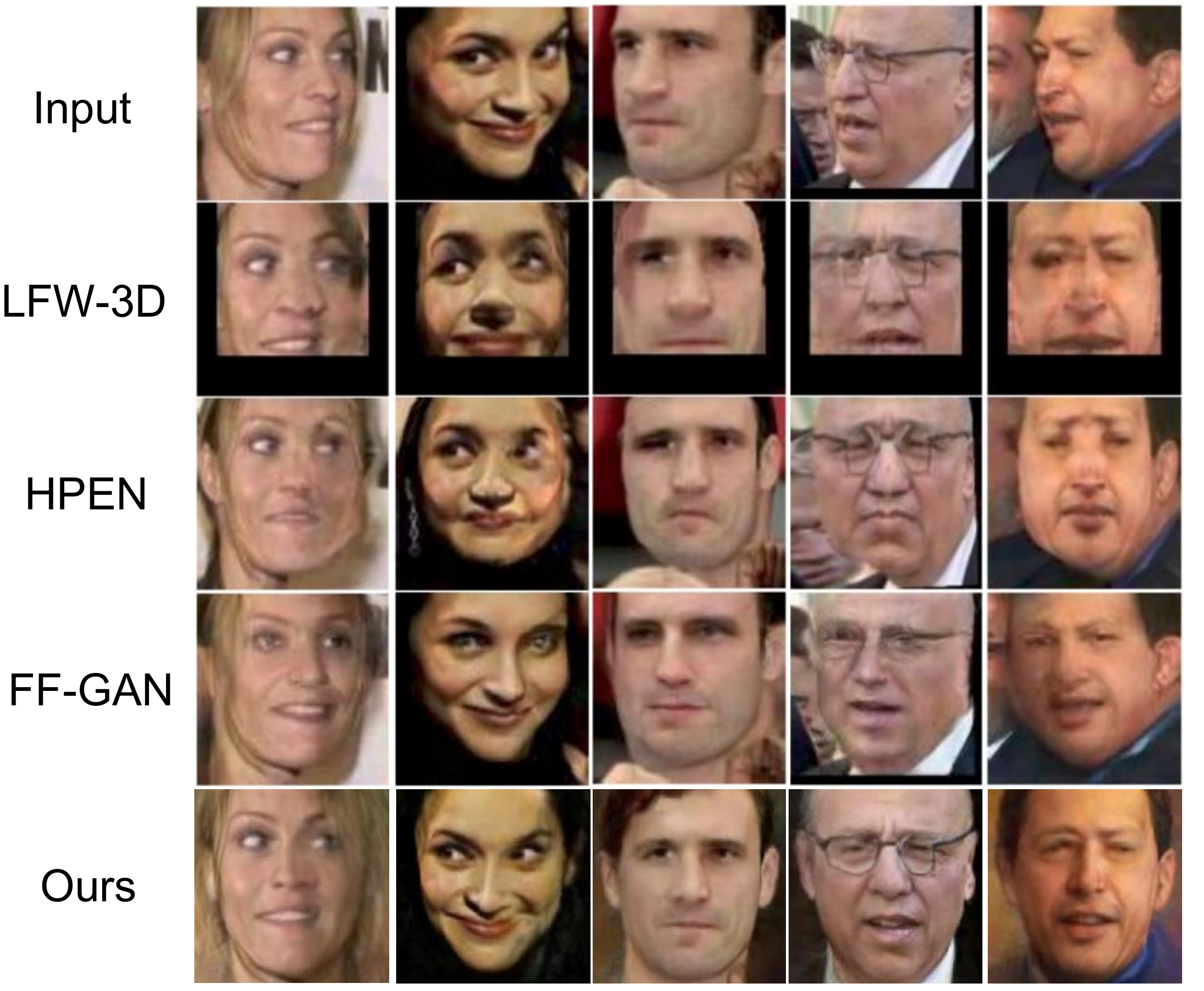}

\caption{Frontalization results with 3D-based Methods. the first row is the input. From top to down rows are results from:  LFW-3D~\cite{hassner2015effective};  HPEN~\cite{zhu2015high};  FF-GAN~\cite{yin2017face} and the last is ours. The samples are selected from LFW~\cite{huang2008labeled}.}

\label{fig:3d}
\vspace{-10pt}
\end{figure}

\subsection{Qualitative Results}
As most papers do not release their code and models, we directly borrow the results reported from some of the papers and perform frontalization with the corresponding image in the reported dataset. Our results are cropped for better visualization. It is recommended to zoom-in for a better view. Results on \textbf{CelebA-HQ}~\cite{karras2017progressive} are shown in Fig~\ref{fig:celeba4} to validate that we can generate high-quality results under extreme poses and expressions

\noindent{\textbf{Comparison with 3D-based methods.}} Fig~\ref{fig:3d} illustrates the results of 3D-based methods and FFGAN~\cite{yin2017face}, which combines 3D and GAN. It can be seen from the figure that while pure 3D methods attempt to fill the missing part with symmetric priors, they would create great artifacts when coming to large poses. This figure is extracted from FFGAN~\cite{yin2017face}. However, FFGAN fails to rotate faces to the front fully with serious loss of details. Our results seem much appealing comparing to theirs.


\begin{figure}[t!]
\centering
\includegraphics[width=1\linewidth]{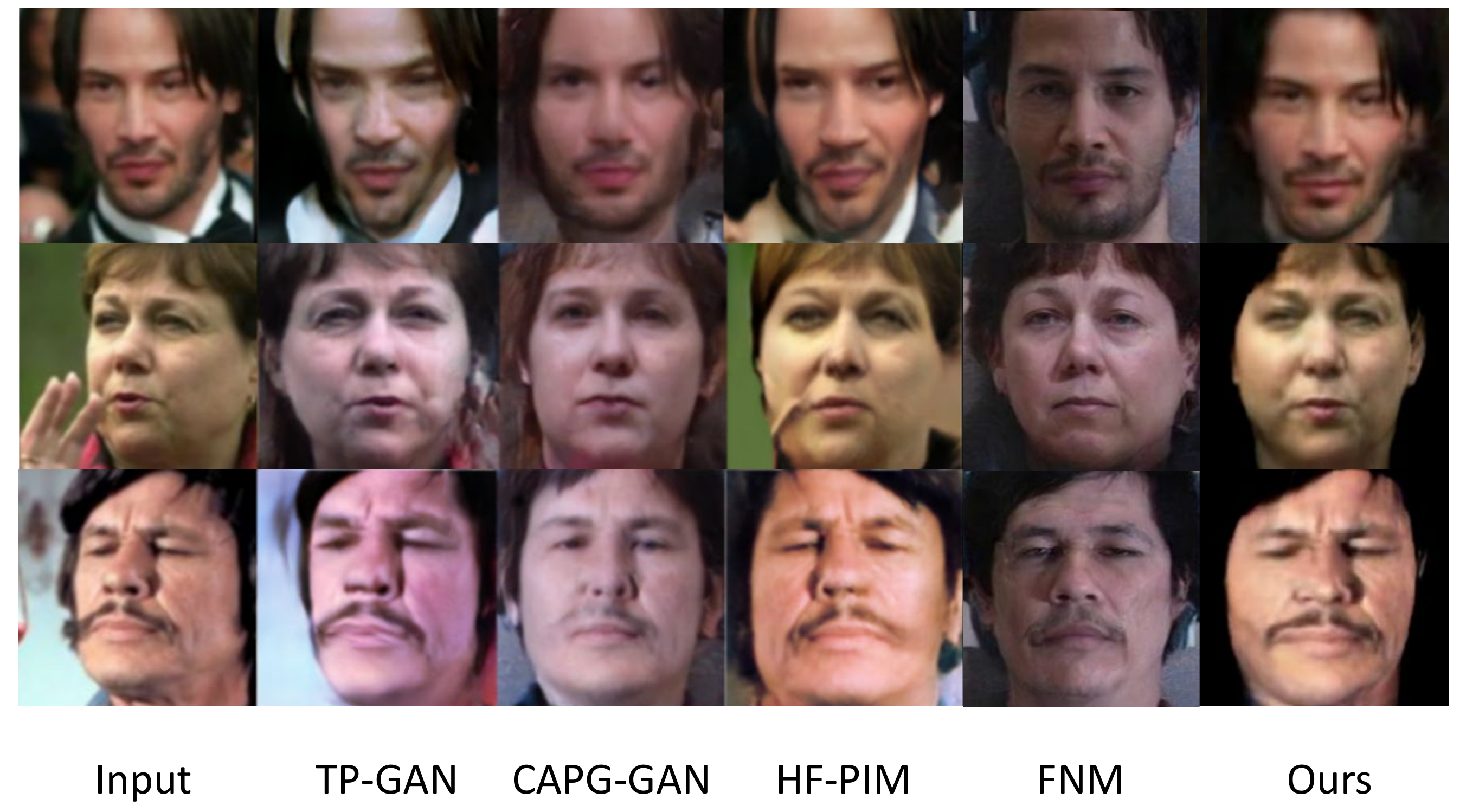}
\caption{Frontalization results comparing with GAN-based methods. The samples are selected from LFW.}
\vspace{-5pt}
\label{fig:gan}
\end{figure}




\noindent{\textbf{Comparison with GAN-based methods.}} Fig~\ref{fig:gan} depicts our comparisons with GAN-based methods. TP-GAN~\cite{huang2017beyond}, CAPG-GAN~\cite{hu2018pose}, and FNM~\cite{qian2019unsupervised} are all purely reconstruction-based methods trained on a constrained dataset. As a consequence, the generated results lie only in the domain where their networks are trained. This limits their applications in the field of entertainment. Ours, on the other  hand, preserves the illumination and environment satisfyingly. Besides, the results of TP-GAN~\cite{huang2017beyond} change the facial shape, CAPG-GAN~\cite{hu2018pose} suffers from identity change. HF-PIM~\cite{cao2018learning} is the previous state-of-the-art method. However, the results reported
seem to change more details of identity than that of ours. For example, nose and jaw shapes on the first person (Keanu Reeves) in their results have been changed. FNM~\cite{qian2019unsupervised} is a GAN-based method that takes both constraint and in-the-wild data. Their results tend to generate the standard frontal face with neutral expressions in even the same background.

\noindent\textbf{Numerical Results.} We have conducted an additional numerical experiments on VidTIMIT dataset~\cite{sanderson2009multi} to directly compare frontalization results on pixel level. The multi-view frames part is used. We select frames with yaw angle larger than $45^{\circ}$ as input and frontal face as ground truth. We unify the scale to 128$\times$128 to compute the PSNR. Only open-sourced methods are used for comparison. The results of ours, FNM~\cite{qian2019unsupervised} and CR-GAN~\cite{tian2018cr} are 28.8, 27.9 and 28.2 (dB), respectively. This is only for reference because it is almost impossible for frontalized results to match the ground truth. However, we can still find our results outperform others.

\subsection{Face Recognition Results}
\noindent{\textbf{Face Recognition Settings.}} Normally, previous works report face recognition results to validate the success of their identity preserving quality after frontalization. However, clear descriptions about their settings are seldom given. Moreover, they report results on the baseline of LightCNN~\cite{wu2018light} with different structures and baseline scores, which are non-comparable. As a result, we will only report their final numbers as a reference.

Meant for proposing a real-world applicable system, we use the state-of-the-art loss function ArcFace~\cite{deng2019arcface} and standard ResNet18~\cite{he2016deep} backbone which has slightly smaller parameter size than LightCNN29~\cite{wu2018light} for face recognition. Both results trained on CASIA and MS1MV2 are provided. We propose a different way of using frontalized data to boost the performance of face recognition by augmenting the original datasets with our 3D generated ones. Then the networks are trained on our augmented datasets. For fair comparison, best results are provided for baselines.


\setlength{\tabcolsep}{4.2pt}
\begin{table}[t] 
\begin{center}

\begin{tabular}{clll}
\hline
Training Data   & Method    & ACC(\%)  & AUC(\%)   \\ \hline
CASIA     & TP-GAN\cite{huang2017beyond}   & 96.13 & 99.42 \\
CASIA     & FF-GAN\cite{yin2017face}   & 96.42 & 99.45 \\
-     & CAPG-GAN\cite{hu2018pose}  & 99.37 & 99.90 \\
-     & LightCNN in \cite{cao2018learning} & 99.39 & 99.87 \\
-     & HF-PIM\cite{cao2018learning}    & 99.41 & 99.92 \\
\hline
CASIA     & Res18     & 98.77 &    99.90   \\
\bf CASIA+rot.  & \bf Res18(ours)     &\bf 98.95 & \bf 99.91     \\
MS1MV2    & Res18     & 99.63 &   99.91    \\
\bf MS1MV2+rot. & \bf Res18(ours)     &\bf 99.68 &  \bf 99.92   \\
\hline 
\end{tabular}
\caption{Verification performance on LFW dataset. Our method can still boost the performance on a strong baseline.}

\label{table:lfw}
\end{center}
\vspace{-10pt}
\end{table}
\setlength{\tabcolsep}{1.4pt}


\setlength{\tabcolsep}{3.5pt}
\begin{table}[t] 
\begin{center}
\begin{tabular}{clll}
\hline
Training Data   & Method & @FAR=.01  & @FAR=.001 \\ \hline

CASIA    & FF-GAN & 85.2$\pm$1.0  & 66.3$\pm$3.3  \\
CASIA     & DR-GAN & 87.2$\pm$1.4  & 78.1$\pm$3.5  \\

CASIA     & FNM\cite{qian2019unsupervised}  & 93.4$\pm$0.9  & 83.8$\pm$2.6  \\
  -   & HF-PIM\cite{cao2018learning} & 95.2$\pm$0.7  & 89.7$\pm$1.4  \\
\hline
CASIA    & Res18 & 90.57$\pm$1.2 & 80.0$\pm$4.1  \\
\bf CASIA+rot.   & \bf Res18(ours)   & \bf 91.98$\pm$0.7 & \bf 82.48$\pm$2.5 \\
MS1MV2    & Res18  & 97.28$\pm$0.6 & 95.39$\pm$0.9 \\
\bf MS1MV2+rot. & \bf Res18(ours)   & \bf 97.30$\pm$0.6  & \bf 95.63$\pm$0.7 \\
\hline 
\end{tabular}
\caption{Verification performance on IJB-A dataset~\cite{klare2015pushing}.}

\label{table:ijb-a}
\end{center}
\vspace{-10pt}
\end{table}
\setlength{\tabcolsep}{1.4pt}


\noindent{\textbf{Quantitative Results.}} We do a comprehensive study on the LFW datatset. The 1:1 face verification accuracy and area-under-curve (AUC) results are reported. It can be discovered that regardless of both weak (CAISA) or strong (MS1M) baselines, augmenting the training set with our frontalized results can boost the performance. Particularly, HF-PIM improves less on their baseline than our proposed method. Improvements can be found on IJB-A as listed in Table~\ref{table:ijb-a} with the verification experiments conducted on ResNet18. MultiPIE~\cite{gross2010multi} is also evaluated with the same settings as~\cite{huang2017beyond} and all works listed in Table \ref{table:multipie}. As stated before, controlled datasets are not our focus. 
However, with our strong baseline, we can improve the recognition results on MultiPIE as well. 

Finally, we conduct experiments on MegaFace dataset, which is wildly used in the field of face recognition. However, no face frontalization paper has reported results on this dataset.  Rank-1 face identification accuracy with 1 million distractors has been evaluated. We use the modified version of the 18 and 50 layers ResNet introduced by the ArcFace~\cite{deng2019arcface}, and compare with their state-of-the-art result on R100.  With our rotated augmentation, we can boost the performance of R50 to outperform the ArcFace R100 model, reaching 98.44\%.

\setlength{\tabcolsep}{4pt}
\begin{table}[t] 
\begin{center}
\begin{tabular}{cccccc}
\hline
Angle & $\pm$30\degree  & $\pm$45\degree & $\pm$60\degree & $\pm$75\degree & $\pm$90\degree\\ \hline
Res18 & 100 & 99.7 & 96.0 & 76.4 &39.0 \\
FF-GAN\cite{yin2017face} & 92.5& 89.7 &  85.2 & 77.2 & 61.2\\
TP-GAN\cite{huang2017beyond} & 98.1 & 95.4 & 87.7 & 77.4 & 64.6 \\
CAPG-GAN\cite{hu2018pose}  & 99.6 & 97.3 & 90.3 & 76.4 & 66.1 \\
HF-PIM\cite{cao2018learning} &  100 & 99.9 & 99.1 & 96.4 & 92.3 \\
\bf Res18(ours)   & \bf 100 & \bf 100 & \bf 99.7 & \bf 99.3 & \bf 94.4
 \\\hline
\end{tabular}
\caption{Rank-1 recognition rates (\%) across views on Multi-PIE dataset.}

\label{table:multipie}
\end{center}
\vspace{-10pt}
\end{table}
\setlength{\tabcolsep}{1.4pt}


\setlength{\tabcolsep}{10pt}
\begin{table}[t] 
\begin{center}
\begin{tabular}{clll}
\hline
Training Data   & Method        & Id\%  \\ \hline
MS1MV2    & R100~(ArcFace) & 98.35 \\
\hline
MS1MV2   & R18         & 97.00   \\
\bf MS1MV2+rotate &\bf  R18         & \bf 97.48 \\
MS1MV2    & R50           & 98.26 \\
\bf MS1MV2+rotate & \bf R50           & \bf 98.44 \\
\hline
\end{tabular}
\caption{Evalution on MegaFace dataset, ``Id'' refers to the rank-1 face identification accuracy with 1M distractors.}

\label{table:megaface}
\end{center}
\vspace{-10pt}
\end{table}
\setlength{\tabcolsep}{1.4pt}



\subsection{Ablation Study}

We train the same network with the same set of parameters and training strategy additionally without (i) the VGG loss; (ii) using multi-scale discriminator and the feature matching loss, as they are proposed together. Instead, we change it to the PatchGAN in the original Pix2Pix paper~\cite{pix2pix2017}. An additional $L_1$ loss on image level is added for regularization. 

\noindent{\textbf{Quantitative Results.}} Frechet Inception Distance (FID)~\cite{heusel2017gans} is widely used to measure the distances between generated and real image domains. 
The results of FID scores are listed in Table~\ref{table:ablation2}. We can observe that the performance reaches the best FID scores with our full model.

\noindent{\textbf{Qualitative Results.}} We show two sets of frontalization results on the LFW dataset. It can be seen that without the multi-scale D, it is difficult for the network to discriminate the real domain, so the results contain certain GAN artifacts. Without the VGG loss, the results tend to be smooth. They are both crucial when it comes to large poses.

Note that the modules we use are the least ones to generate reasonable results. Our method can still create visible artifacts, but there is great potential for improvements. 

\begin{figure}[t!]
\centering

\vspace{0pt}
\includegraphics[width=1\linewidth]{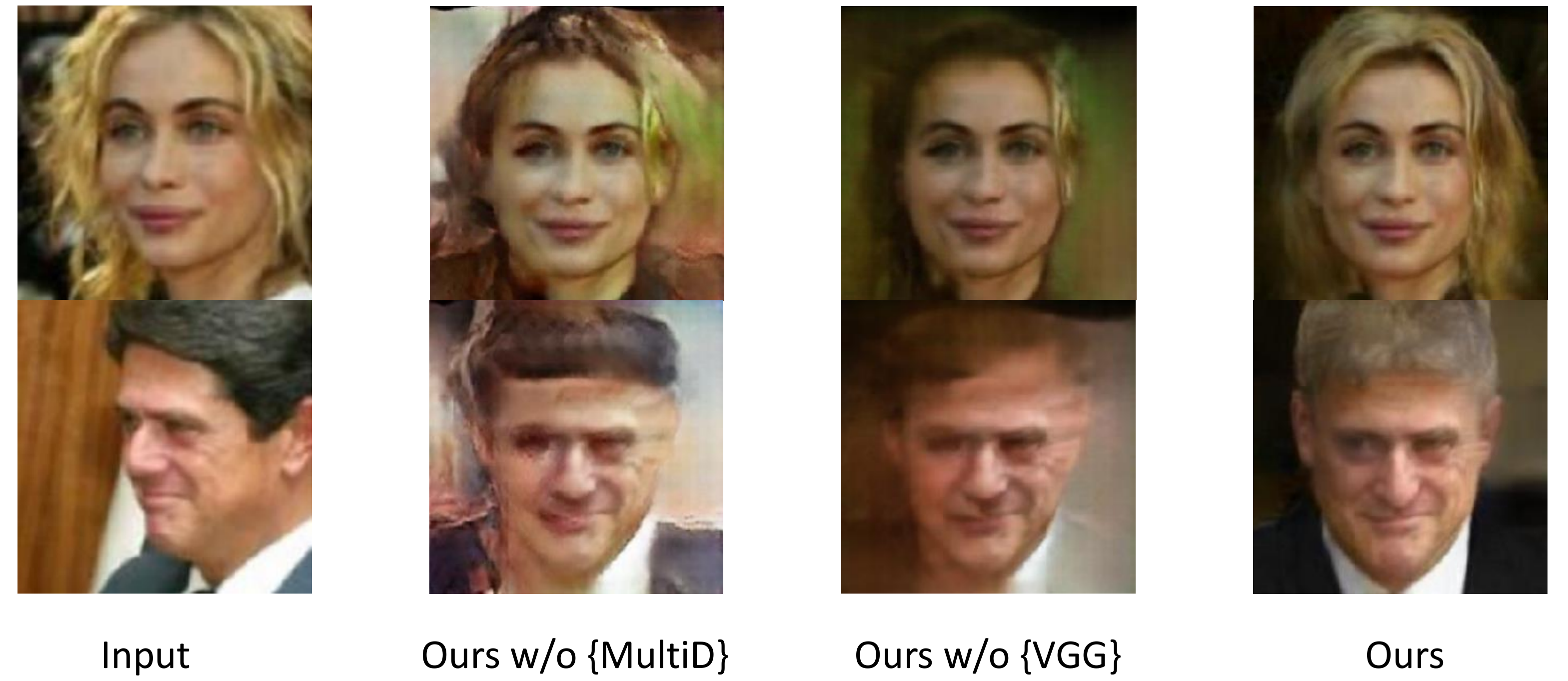}

\caption{Ablation study of frontalization results.}
\vspace{-10pt}
\label{fig:abltaion}
\end{figure}


\setlength{\tabcolsep}{4pt}
\begin{table}[t] 
\begin{center}

\begin{tabular}{ccc}
\hline

~~~~~~Approach  $\setminus$ Dataset~~~~~~ & LFW & ~~~~~~IJB-A~~~~~~ \\
\noalign{\smallskip}

\hline
~~~~~~Ours w/o \{VGG\}~~~~~~ & 93.1 & ~~~~~~126.5~~~~~~ \\
~~~~~~Ours w/o \{MultiD\}~~~~~~  &  83.9 & ~~~~~~132.3~~~~~~  \\
~~~~~~\textbf{Ours}~~~~~~&  \textbf{83.1} & ~~~~~~\textbf{70.9}~~~~~~ \\
\hline
\end{tabular}
\caption{Ablation study on loss functions with FID metric. For FID scores, the lower the better.}

\label{table:ablation2}
\end{center}
\vspace{-10pt}
\end{table}
\setlength{\tabcolsep}{1.4pt}

\section{Conclusion}

In this work, we take the advantage of 3D face priors and propose a novel strategy called Rotate-and-Render for unsupervised face rotation. Our key insight is to create self-supervision signal by rotating and rendering the roughly-predicted 3D representation to a random pose and back to its original place. So that self-supervision can be leveraged to create photo-realistic results by translating rendered image to real-image domain. 
Through comprehensive experiments, the following strengths of our pipeline have been validated: {\bf 1)} No multi-view or paired-data, nor any kind of labels, including identities are needed for training our method.
{\bf 2)} Instead of purely frontalization a single-view face can be rotated to any desired angle. {\bf 3)} Near photorealistic face rotation results which preserve details and illumination condition can be generated.  Visualization indicates the superiority of our method. {\bf 4)} It can be used to augment face datasets and boost recognition results on large-scale benchmarks with strong baseline models. 

{\small
\noindent\textbf{Acknowledgements.}
We thank Hao Shao for helpful discussions. This work is supported in part by SenseTime Group Limited, and in part by the General Research Fund through the Research Grants Council of Hong Kong under Grants CUHK14202217, CUHK14203118, CUHK14205615, CUHK14207814, CUHK14208619, CUHK14213616.
}


{\small
\bibliographystyle{ieee_fullname}
\bibliography{egbib}

\begin{thebibliography}{10}\itemsep=-1pt

\bibitem{blanz1999morphable}
Volker Blanz, Thomas Vetter, et~al.
\newblock A morphable model for the synthesis of 3d faces.
\newblock In {\em Siggraph}, 1999.

\bibitem{cao2018learning}
Jie Cao, Yibo Hu, Hongwen Zhang, Ran He, and Zhenan Sun.
\newblock Learning a high fidelity pose invariant model for high-resolution
  face frontalization.
\newblock In {\em NeurIPS}, 2018.

\bibitem{deng2018uv}
Jiankang Deng, Shiyang Cheng, Niannan Xue, Yuxiang Zhou, and Stefanos
  Zafeiriou.
\newblock Uv-gan: Adversarial facial uv map completion for pose-invariant face
  recognition.
\newblock In {\em CVPR}, 2018.

\bibitem{deng2019arcface}
Jiankang Deng, Jia Guo, Niannan Xue, and Stefanos Zafeiriou.
\newblock Arcface: Additive angular margin loss for deep face recognition.
\newblock In {\em CVPR}, 2019.

\bibitem{ge2018fd}
Yixiao Ge, Zhuowan Li, Haiyu Zhao, Guojun Yin, Shuai Yi, Xiaogang Wang, et~al.
\newblock Fd-gan: Pose-guided feature distilling gan for robust person
  re-identification.
\newblock In {\em Advances in neural information processing systems}, pages
  1222--1233, 2018.

\bibitem{goodfellow2014generative}
Ian~J. {Goodfellow}, Jean {Pouget-Abadie}, Mehdi {Mirza}, Bing {Xu}, David
  {Warde-Farley}, Sherjil {Ozair}, Aaron~C. {Courville}, and Yoshua {Bengio}.
\newblock Generative adversarial nets.
\newblock In {\em NeurIPS}, 2014.

\bibitem{gross2010multi}
Ralph Gross, Iain Matthews, Jeffrey Cohn, Takeo Kanade, and Simon Baker.
\newblock Multi-pie.
\newblock {\em Image and Vision Computing}.

\bibitem{guo2016ms}
Yandong Guo, Lei Zhang, Yuxiao Hu, Xiaodong He, and Jianfeng Gao.
\newblock Ms-celeb-1m: A dataset and benchmark for large-scale face
  recognition.
\newblock In {\em ECCV}, 2016.

\bibitem{hassner2015effective}
Tal Hassner, Shai Harel, Eran Paz, and Roee Enbar.
\newblock Effective face frontalization in unconstrained images.
\newblock In {\em CVPR}, 2015.

\bibitem{he2016deep}
Kaiming He, Xiangyu Zhang, Shaoqing Ren, and Jian Sun.
\newblock Deep residual learning for image recognition.
\newblock In {\em CVPR}, 2016.

\bibitem{heusel2017gans}
Martin Heusel, Hubert Ramsauer, Thomas Unterthiner, Bernhard Nessler, and Sepp
  Hochreiter.
\newblock Gans trained by a two time-scale update rule converge to a local nash
  equilibrium.
\newblock In {\em NeurIPS}, 2017.

\bibitem{hu2018pose}
Yibo Hu, Xiang Wu, Bing Yu, Ran He, and Zhenan Sun.
\newblock Pose-guided photorealistic face rotation.
\newblock In {\em CVPR}, 2018.

\bibitem{huang2008labeled}
Gary~B Huang, Marwan Mattar, Tamara Berg, and Eric Learned-Miller.
\newblock Labeled faces in the wild: A database forstudying face recognition in
  unconstrained environments.
\newblock 2008.

\bibitem{huang2017beyond}
Rui Huang, Shu Zhang, Tianyu Li, and Ran He.
\newblock Beyond face rotation: Global and local perception gan for
  photorealistic and identity preserving frontal view synthesis.
\newblock In {\em ICCV}, 2017.

\bibitem{ioffe2015batch}
Sergey Ioffe and Christian Szegedy.
\newblock Batch normalization: Accelerating deep network training by reducing
  internal covariate shift.
\newblock {\em arXiv preprint arXiv:1502.03167}, 2015.

\bibitem{pix2pix2017}
Phillip Isola, Jun-Yan Zhu, Tinghui Zhou, and Alexei~A Efros.
\newblock Image-to-image translation with conditional adversarial networks.
\newblock {\em CVPR}, 2017.

\bibitem{karras2017progressive}
Tero Karras, Timo Aila, Samuli Laine, and Jaakko Lehtinen.
\newblock Progressive growing of gans for improved quality, stability, and
  variation.
\newblock {\em ICLR}, 2018.

\bibitem{kato2018neural}
Hiroharu Kato, Yoshitaka Ushiku, and Tatsuya Harada.
\newblock Neural 3d mesh renderer.
\newblock In {\em CVPR}, 2018.

\bibitem{klare2015pushing}
Brendan~F Klare, Ben Klein, Emma Taborsky, Austin Blanton, Jordan Cheney,
  Kristen Allen, Patrick Grother, Alan Mah, and Anil~K Jain.
\newblock Pushing the frontiers of unconstrained face detection and
  recognition: Iarpa janus benchmark a.
\newblock In {\em CVPR}, 2015.

\bibitem{lee2019maskgan}
Cheng-Han Lee, Ziwei Liu, Lingyun Wu, and Ping Luo.
\newblock Maskgan: towards diverse and interactive facial image manipulation.
\newblock In {\em Proceedings of the IEEE Conference on Computer Vision and
  Pattern Recognition}, 2020.

\bibitem{liu2019learning}
Xihui Liu, Guojun Yin, Jing Shao, Xiaogang Wang, and Hongsheng Li.
\newblock Learning to predict layout-to-image conditional convolutions for
  semantic image synthesis.
\newblock In {\em Advances in Neural Information Processing Systems}, pages
  568--578, 2019.

\bibitem{liu2015deep}
Ziwei Liu, Ping Luo, Xiaogang Wang, and Xiaoou Tang.
\newblock Deep learning face attributes in the wild.
\newblock In {\em ICCV}, 2015.

\bibitem{miyato2018spectral}
Takeru Miyato, Toshiki Kataoka, Masanori Koyama, and Yuichi Yoshida.
\newblock Spectral normalization for generative adversarial networks.
\newblock {\em arXiv preprint arXiv:1802.05957}, 2018.

\bibitem{moniz2018unsupervised}
Joel Ruben~Antony Moniz, Christopher Beckham, Simon Rajotte, Sina Honari, and
  Chris Pal.
\newblock Unsupervised depth estimation, 3d face rotation and replacement.
\newblock In {\em Advances in Neural Information Processing Systems}, pages
  9736--9746, 2018.

\bibitem{nech2017level}
Aaron Nech and Ira Kemelmacher-Shlizerman.
\newblock Level playing field for million scale face recognition.
\newblock In {\em CVPR}, 2017.

\bibitem{park2019semantic}
Taesung Park, Ming-Yu Liu, Ting-Chun Wang, and Jun-Yan Zhu.
\newblock Semantic image synthesis with spatially-adaptive normalization.
\newblock In {\em CVPR}, 2019.

\bibitem{pathak2016context}
Deepak Pathak, Philipp Krahenbuhl, Jeff Donahue, Trevor Darrell, and Alexei~A
  Efros.
\newblock Context encoders: Feature learning by inpainting.
\newblock In {\em Proceedings of the IEEE Conference on Computer Vision and
  Pattern Recognition}, pages 2536--2544, 2016.

\bibitem{pumarola2018ganimation}
Albert Pumarola, Antonio Agudo, Aleix~M Martinez, Alberto Sanfeliu, and
  Francesc Moreno-Noguer.
\newblock Ganimation: Anatomically-aware facial animation from a single image.
\newblock In {\em Proceedings of the European Conference on Computer Vision
  (ECCV)}, pages 818--833, 2018.

\bibitem{qian2019make}
Shengju Qian, Kwan-Yee Lin, Wayne Wu, Yangxiaokang Liu, Quan Wang, Fumin Shen,
  Chen Qian, and Ran He.
\newblock Make a face: Towards arbitrary high fidelity face manipulation.
\newblock In {\em Proceedings of the IEEE International Conference on Computer
  Vision}, pages 10033--10042, 2019.

\bibitem{qian2019unsupervised}
Yichen Qian, Weihong Deng, and Jiani Hu.
\newblock Unsupervised face normalization with extreme pose and expression in
  the wild.
\newblock In {\em CVPR}, 2019.

\bibitem{sanderson2009multi}
Conrad Sanderson and Brian~C Lovell.
\newblock Multi-region probabilistic histograms for robust and scalable
  identity inference.
\newblock In {\em International conference on biometrics}, pages 199--208.
  Springer, 2009.

\bibitem{tian2018cr}
Yu Tian, Xi Peng, Long Zhao, Shaoting Zhang, and Dimitris~N Metaxas.
\newblock Cr-gan: learning complete representations for multi-view generation.
\newblock {\em IJCAI}, 2018.

\bibitem{tran2017disentangled}
Luan Tran, Xi Yin, and Xiaoming Liu.
\newblock Disentangled representation learning gan for pose-invariant face
  recognition.
\newblock In {\em CVPR}, 2017.

\bibitem{tran2018representation}
Luan~Quoc Tran, Xi Yin, and Xiaoming Liu.
\newblock Representation learning by rotating your faces.
\newblock {\em IEEE transactions on pattern analysis and machine intelligence},
  2018.

\bibitem{wang2018high}
Ting-Chun Wang, Ming-Yu Liu, Jun-Yan Zhu, Andrew Tao, Jan Kautz, and Bryan
  Catanzaro.
\newblock High-resolution image synthesis and semantic manipulation with
  conditional gans.
\newblock In {\em CVPR}, 2018.

\bibitem{wang2019learning}
Xiaolong Wang, Allan Jabri, and Alexei~A Efros.
\newblock Learning correspondence from the cycle-consistency of time.
\newblock In {\em Proceedings of the IEEE Conference on Computer Vision and
  Pattern Recognition}, pages 2566--2576, 2019.

\bibitem{wang2018image}
Yi Wang, Xin Tao, Xiaojuan Qi, Xiaoyong Shen, and Jiaya Jia.
\newblock Image inpainting via generative multi-column convolutional neural
  networks.
\newblock In {\em Advances in neural information processing systems}, 2018.

\bibitem{wayne2018reenactgan}
Wayne Wu, Yunxuan Zhang, Cheng Li, Chen Qian, and Chen~Change Loy.
\newblock Reenactgan: Learning to reenact faces via boundary transfer.
\newblock In {\em ECCV}, 2018.

\bibitem{wu2018light}
Xiang Wu, Ran He, Zhenan Sun, and Tieniu Tan.
\newblock A light cnn for deep face representation with noisy labels.
\newblock {\em IEEE Transactions on Information Forensics and Security}, 2018.

\bibitem{yeh2017semantic}
Raymond~A Yeh, Chen Chen, Teck Yian~Lim, Alexander~G Schwing, Mark
  Hasegawa-Johnson, and Minh~N Do.
\newblock Semantic image inpainting with deep generative models.
\newblock In {\em Proceedings of the IEEE Conference on Computer Vision and
  Pattern Recognition}, pages 5485--5493, 2017.

\bibitem{yi2014learning}
Dong Yi, Zhen Lei, Shengcai Liao, and Stan~Z Li.
\newblock Learning face representation from scratch.
\newblock {\em arXiv preprint arXiv:1411.7923}, 2014.

\bibitem{yim2015rotating}
Junho Yim, Heechul Jung, ByungIn Yoo, Changkyu Choi, Dusik Park, and Junmo Kim.
\newblock Rotating your face using multi-task deep neural network.
\newblock In {\em CVPR}, 2015.

\bibitem{yin2019instance}
Weidong Yin, Ziwei Liu, and Chen~Change Loy.
\newblock Instance-level facial attributes transfer with geometry-aware flow.
\newblock In {\em AAAI}, 2019.

\bibitem{yin2017face}
Xi Yin, Xiang Yu, Kihyuk Sohn, Xiaoming Liu, and Manmohan Chandraker.
\newblock Towards large-pose face frontalization in the wild.
\newblock In {\em ICCV}, 2017.

\bibitem{zhou2020rotate}
Hang Zhou, Jihao Liu, Ziwei Liu, Yu Liu, and Xiaogang Wang.
\newblock Rotate-and-render: Unsupervised photorealistic face rotation from
  single-view images.
\newblock In {\em IEEE Conference on Computer Vision and Pattern Recognition
  (CVPR)}, 2020.

\bibitem{zhou2019talking}
Hang Zhou, Yu Liu, Ziwei Liu, Ping Luo, and Xiaogang Wang.
\newblock Talking face generation by adversarially disentangled audio-visual
  representation.
\newblock In {\em AAAI}, 2019.

\bibitem{zhou2019vision}
Hang Zhou, Ziwei Liu, Xudong Xu, Ping Luo, and Xiaogang Wang.
\newblock Vision-infused deep audio inpainting.
\newblock In {\em ICCV}, 2019.

\bibitem{CycleGAN2017}
Jun-Yan Zhu, Taesung Park, Phillip Isola, and Alexei~A Efros.
\newblock Unpaired image-to-image translation using cycle-consistent
  adversarial networkss.
\newblock In {\em Computer Vision (ICCV), 2017 IEEE International Conference
  on}, 2017.

\bibitem{zhu2015high}
Xiangyu Zhu, Zhen Lei, Junjie Yan, Dong Yi, and Stan~Z Li.
\newblock High-fidelity pose and expression normalization for face recognition
  in the wild.
\newblock In {\em CVPR}, 2015.

\bibitem{zhu2017face}
Xiangyu Zhu, Xiaoming Liu, Zhen Lei, and Stan~Z Li.
\newblock Face alignment in full pose range: A 3d total solution.
\newblock {\em IEEE transactions on pattern analysis and machine intelligence},
  41(1):78--92, 2017.

\end{thebibliography}
}


\begin{figure*}[t!]
\centering
\includegraphics[width=\linewidth]{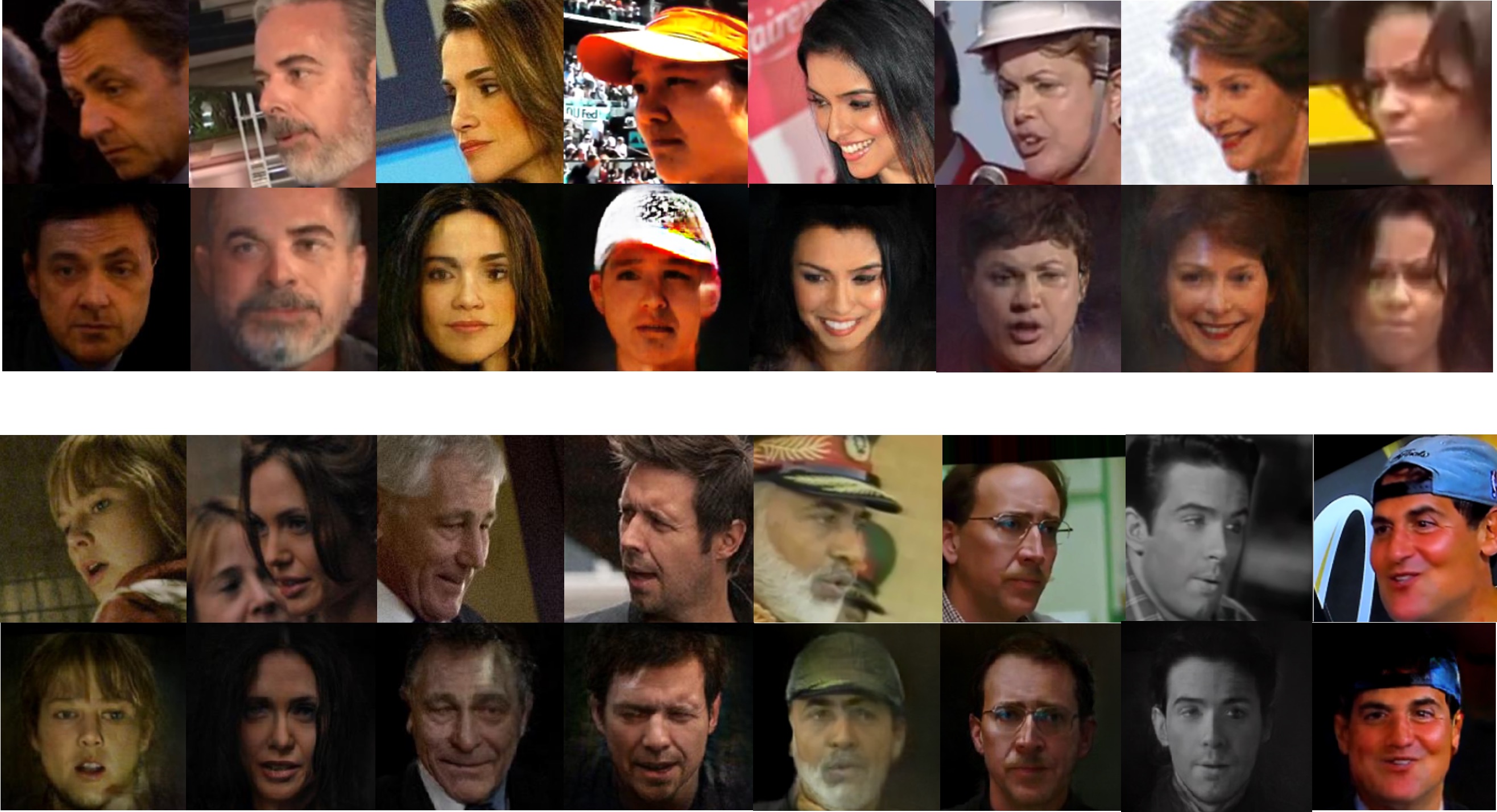}
\caption{Additional frontalization result. The first rows are the inputs and the second rows are the rotated results.
 }
\vspace{-5pt}
\label{fig:result}
\end{figure*}


\newpage

\appendix
\addcontentsline{toc}{section}{Appendices}

\section*{Appendices}

\subsection*{A. Possibility for Improvements}

In our paper~\cite{zhou2020rotate}, we describe the least building blocks for our proposed method to work. However, there is actually great space for improvements. For example, the 3DDFA~\cite{zhu2017face} model used for 3D face fitting is quite inaccurate, thus can be replaced by any state-of-the-art model. Moreover, the Render-to-Image network could be replaced with the one in Pix2pix HD~\cite{wang2018high} or modified to a progressive growing style~\cite{karras2017progressive}. Upgraded discriminator structures and training losses are also applicable.

Here we propose a simple modification to our Render-to-Image network. The idea is to use facial semantic maps to guide the generation procedure, which shares substantially similar idea with SPADE~\cite{park2019semantic} and MaskGAN~\cite{lee2019maskgan}. Semantic maps are normally acquired by an additional face parsing model. Differently, we leverage facial landmarks that can be directly derived from 3D models. Then we connect and expend the landmarks to get rough predictions of the key components' semantic maps. The predicted maps are converted to weights and biases of the batch normalization parameters inside the ResNet blocks of our original generator $\text{G}$. Please refer to our released code \footnote{\url{https://github.com/Hangz-nju-cuhk/Rotate-and-Render}.} for our implementation and more details.



\subsection*{B. Additional Results}

We show more frontalization results in the figure~\ref{fig:result} to validate that our method can work robustly with different poses, illuminating conditions and qualities. 

\end{document}